\documentclass[sigconf]{acmart}
\acmConference[ISSTA 2024]{ACM SIGSOFT International Symposium on Software Testing and Analysis}{16-20 September, 2024}{Vienna, Austria}
\usepackage[utf8]{inputenc}
\usepackage{enumerate}
\usepackage{amsfonts}
\usepackage{amsmath}
\usepackage{graphicx} 
\usepackage[]{color}
\usepackage{amsthm}
\usepackage{multirow}
\newtheorem{definition}{Definition}
\usepackage{tcolorbox}
\usepackage[ruled,vlined, noend]{algorithm2e}

\newtcolorbox{graybox}{
  colback=gray!20,
  colframe=black,
  boxrule=1pt, 
  arc=3mm,     
  boxsep=0.5pt,
  after={\vspace{-5mm}}
}

\newtcolorbox{whitebox}{
  colback=white,
  colframe=black,
  boxrule=1pt, 
  arc=5mm,     
  boxsep=1pt,
}

\definecolor{cb_orange}{rgb}{1.0,0.51,0.0}
\definecolor{cb_blue}{rgb}{0.22,0.49,0.72}
\definecolor{cb_green}{rgb}{0.3,0.67,0.29}
\definecolor{cb_red}{rgb}{0.89,0.1,0.11}
\definecolor{cb_purple}{rgb}{0.6, 0.31, 0.64}
\definecolor{cb_pink}{rgb}{0.858, 0.188, 0.478}
\newcommand{\dat}[1]{{#1}}

\newcounter{todocounter}

\usepackage{xspace}
\newcommand{\tool}{\textsc{VRDSynth}\xspace}
\SetAlgoSkip{}

\title{VRDSynth: Synthesizing Programs for Multilingual Visually Rich Document Information Extraction}
\author{Dat Nguyen}
\affiliation{
  \institution{University of Melbourne}
  \city{Melbourne}
  \state{VIC}
  \country{Australia}
}
\email{thanhdatn@student.unimelb.edu.au}

\author{Tung Do-Viet}
\affiliation{
  \institution{Cinnamon AI}
  \city{Ho Chi Minh City}
  \country{Vietnam}
}
\email{ace@cinnamon.is}

\author{Hung Nguyen-Duy}
\affiliation{
  \institution{Independent Researcher}
  \city{Hanoi}
  \country{Vietnam}
}
\email{hungnd1226@gmail.com}

\author{Tuan-Hai Luu}
\affiliation{
  \institution{Cinnamon AI}
  \city{Ho Chi Minh City}
  \country{Vietnam}
}
\email{sam@cinnamon.is}

\author{Hung Le}
\affiliation{
  \institution{Deakin University}
  \city{Geelong}
  \state{VIC}
  \country{Australia}
}
\email{thai.le@deakin.edu.au}

\author{Bach Le}
\affiliation{
  \institution{University of Melbourne}
  \city{Melbourne}
  \state{VIC}
  \country{Australia}
}
\email{bach.le@unimelb.edu.au}

\author{Patanamon (Pick) Thongtanunam}
\affiliation{
  \institution{University of Melbourne}
  \city{Melbourne}
  \state{VIC}
  \country{Australia}
}
\email{patanamon.t@unimelb.edu.au}
\date{July. 2024}

\begin{document}

\begin{abstract}

Businesses often need to query visually rich documents (VRDs), e.g., purchase receipts, medical records, and insurance forms, among many other forms from multiple vendors, to make informed decisions. As such, several techniques have been proposed to automatically extract independent entities of interest from VRDs such as extracting price tags from purchase receipts, etc. However, for extracting semantically linked entities, such as finding corresponding price tags for each item, these techniques either have limited capability in handling new layouts, e.g., template-based approaches, or require extensive amounts of pre-training data and do not perform well, e.g., deep-learning approaches.



In this work, we introduce a program synthesis method, namely \tool{}, to automatically generate programs to extract entity relations from multilingual VRDs. 
Two key novelties, which empower \tool{} to tackle flexible layouts while requiring no pre-training data for extracting entity relations, include: (1) a new domain-specific language (DSL) to effectively capture the spatial and textual relations between document entities, and (2) a novel synthesis algorithm that makes use of frequent spatial relations between entities to construct initial programs, equivalent reduction to prune the search space, and a combination of positive, negative, and mutually exclusive programs to improve the coverage of programs.

We experiment on two popular VRD understanding benchmarks, namely FUNSD and XFUND, on the semantic entity linking task, consisting of 1,592 forms in 8 different languages. \tool{}, despite having no prior pre-training data, outperforms the state-of-the-art pre-trained deep-learning approaches - LayoutXLM, \dat{$\text{InfoXLM}_{Base}$ and $\text{XLMRoberta}_{Base}$ in 5, 6 and 7 out of 8 languages, respectively}. Noticeably, in English, \tool{} improves 42\% of F1 score over LayoutXLM while being complementary to LayoutXLM in 7/8 languages.
\dat{We further extend $\tool{}$ with automated table recognition, namely $\tool{}_{Table}$, and compare with $\text{InfoXLM}_{Large}$ and $\text{XLMRoberta}_{Large}$. $\tool{}_{Table}$ outperforms these baselines on 4 out of 8 languages and in terms of average F1 score.}
\tool{} significantly improves the memory footprint required for program storage and inference over \dat{all baselines} (1M and 380MB versus that of 1.48GB and 3GB required by LayoutXLM), while maintaining similar time efficiency despite the speed differences between the languages used for implementation (Python vs C++). 
\end{abstract}
\maketitle

\vspace{-8pt}

\section{Introduction}

Businesses often have to collect and store information from various sources for administrative tasks: from medical records, bills, purchase receipts, and insurance forms to technical reports from different vendors.
These administrative documents often have various layouts with different visual elements such as images, graphics, tables, diagrams, etc, and hence, they are also called visually-rich documents (VRDs)~\cite{Iyer}.
A massive amount of data today is stored in the form of VRDs as a way for business enterprises to retain and exchange information. Often, businesses need to query this data to make informed decisions, e.g., enterprises need to extract prices for sold items from purchase receipts to claim credits for goods and services tax (GST). This motivates several recent approaches that can automatically extract useful information from VRDs~\cite{xu2019layoutlm, xu2021layoutlmv2, huang2022layoutlmv3, Liu2019, Lohani2019, Sunder2019, Madashi2019, dang2020msaupaf, Katti2018, Palm2018, doctemplate_comp_blocks_HanchuanPeng2000,doctemplate_comp_block_prj_Peng2001}.

\begin{figure*}
    \centering
    \includegraphics[width=\textwidth]{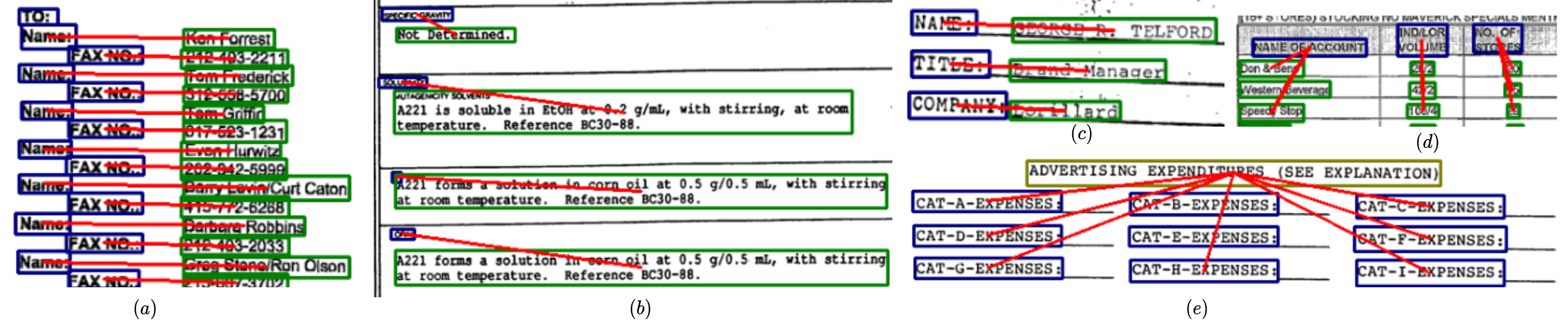}
    \caption{Example output of \tool{}'s synthesized programs in different layout: (a) List of key-value group, (b) key and long value, (c) list of key-value (flat) (d) table and (e) header and key-value table. The programs collectively link each corresponding key (e.g., names, fax) to corresponding values or headers to corresponding keys. Note that we show the original images to also demonstrate the problem of varying quality of scanned VRDs.}
    \label{fig:middle_representation}
\end{figure*}

Information extraction from VRDs consists of two phases: recognition of semantic entities and extraction of relations between entities~\cite{Jaume2019}. 
Semantic entity recognition consists of identifying which entity belongs to which category while relation extraction identifies which value belongs to which key. 
As an example, Figure~\ref{fig:middle_representation} depicts the joint output of the two tasks: given the set of entities (text and corresponding bounding box), semantic entity recognition identifies which entity is header, key, or value (annotated in yellow, blue and green bounding boxes, respectively), while relation extraction links corresponding keys and value together. Both steps are necessary to organize the extracted information. 
While the distinctive layouts and visual features provide sufficient information for state-of-the-art techniques such as~\cite{huang2022layoutlmv3, Xu2020LayoutXLMMP, Liu2019, xu2019layoutlm, xu2021layoutlmv2, huang2022layoutlmv3, zhang2023reading} to perform reasonably in semantic entity recognition, semantic entity linking between these key-and-values is still largely unaddressed~\cite{Jaume2019}.
On FUNSD~\cite{Jaume2019}, BERT-based entity linking approach~\cite{Jaume2019}, which uses only textual features, achieved an F1 score, recall, and precision of only 0.04, 0.99, and 0.02. This means that the approach can identify most of the links, but the majority of them are false positives.
On the multilingual dataset, a state-of-the-art technique based on a large language model, namely, LayoutXLM~\cite{Xu2020LayoutXLMMP} performs reasonably well on key-value linking (0.6 to 0.7 for F1 score) but requires an extensive amount of training data and memory footprint for storage and inference ~\cite{Xu2020LayoutXLMMP, Rogge2023}. Furthermore, when reproduced in our experiments on header-key linking, LayoutXLM performs worse.  

On the other hand, program synthesis techniques, which make use of domain-specific constraints provided by human experts, are capable of synthesizing highly precise programs in data wrangling and information extraction, given a limited amount of data~\cite{Gulwani2017, Gulwani2011FlashFill, Nye2019a, Iyer}.
However, the VRD domain is different from the existing synthesis works that operate on spreadsheets~\cite{Gulwani2011FlashFill, Gulwani2017, Cambronero2023}, sequence of text~\cite{Iyer}, or semi-structured web pages or pre-defined forms~\cite{Parthasarathy2022lrsyn}. 
In general, VRDs are often scanned, e.g., scanned purchase receipts, and thus, they are often noisy (see Figure~\ref{fig:middle_representation}) due to the varying quality of the scanned versions. Because of this, VRDs are considered much more challenging than the types of data that program synthesis techniques can handle as mentioned above.

In this paper, we aim to develop a program synthesis technique, namely, \tool{}, to automatically extract semantic links from entities of interest from VRDs. There are two key novelties that empower \tool{} to tackle flexible layouts while requiring no pre-training data in this entity-relation extraction task, including (1) a new domain-specific language (DSL) to effectively capture the spatial and textual relations between document entities, and
(2) a novel synthesis algorithm that efficiently and effectively navigates through the search space to find high-quality candidate programs. To achieve this, we represent each visually rich document as a collection of entities and spatial relations between them. Each entity consists of a bounding box location, the entity's label (e.g., header, question, answer), and its textual content. Note that the entity label can be obtained by any entity recognizer, e.g.,~ \cite{Xu2020LayoutXLMMP, Qasim2019, huang2022layoutlmv3}. Next, spatial entity relations are built based on spatial alignment (top, down, left, right) between the entities. We propose an expressive DSL, which is a context-free grammar, that can effectively represent these spatial and textual relations between entities while maintaining the tractability of searching for programs that correctly locate all pairs of linkable entities.

To synthesize such programs, we construct the specifications that describe examples of pairs of related and unrelated entities obtained from the training data, for which we use FUNSD and XFUND~\cite{Jaume2019, xu-etal-2022-xfund}. We note that this training data is different from the pre-training data used by LayoutXLM for representation learning~\cite{Xu2020LayoutXLMMP}. While both LayoutXLM and \tool{} require a training dataset, LayoutXLM also requires an additional pre-training dataset, consisting of a large number of forms annotated with bounding box and textual content but without either semantic entity recognition or relation extraction labeling for representation learning.

Finally, we propose a novel program synthesis algorithm for \tool{} to synthesize programs consistent with the DSL and the example-based specifications extracted via the training data in the previous step. The algorithm works in two phases: identifying initial programs and iteratively refining programs. In the first stage, \tool{} mines the frequent relations between semantically linked entities from the training documents and uses these frequent relations to construct initial programs that filter out potential pairs of semantically linkable entities. Since these programs can still be imprecise, \tool{} further refines these programs by searching for textual, label, position, or relation constraints that result in more precise programs. During this search, \tool{} makes use of equivalent reduction~\cite{smith2019equivalentreduction} to prune the search space. At the end of each iterative refinement step, \tool{} combines positive, negative, and mutually exclusive programs to produce more precise programs.

We evaluate \tool{} on the semantic entity linking benchmark from FUNSD and XFUND~\cite{Jaume2019, xu-etal-2022-xfund}. These two datasets together consist of 1,592 forms in 8 languages - English, German, French, Italian, Japanese, Portuguese, and Chinese. On this benchmark, we show that even when requiring no prior pre-training data, \tool{} synthesized programs that outperform LayoutXLM~\cite{Xu2020LayoutXLMMP}\dat{, $\text{InfoXLM}_{Base}$~\cite{chi-etal-2021-infoxlm} and $~\text{XLMRoberta}_{Base}$~\cite{conneau-etal-2020-unsupervised} in terms of F1 score on 5, 6, and 7 out of 8 languages, respectively. In detail, these programs have consistently higher precision than the all pre-trained models (0.732 versus 0.534 of the best average precision baseline-LayoutXLM)}. 
Notably, in English, the F1 score of \tool{} is 42\% higher than the the best baseline - LayoutXLM. 
\dat{To evaluate the extensibility of \tool{} for table relations, we further extended \tool{} with automated table recognition module TATR~\cite{smock2021tabletransformer}, resulting in $\tool{}_{Table}$. We compare $\tool{}_{Table}$ with $\text{InfoXLM}_{Large}$ and $\text{XLMRoberta}_{Large}$, which are the extensions of the base models. Experiment results show that $\tool{}_{Table}$ outperforms both baselines in 4 out of 8 languages and in terms of average F1 score.}
In terms of efficiency, \tool{} requires significantly \dat{less memory than all baselines to store and for inference. } to store (1MB vs 1.48GB) and to run (380MB vs 3.04GB).  \tool{} achieve similar time efficiency to LayoutXLM despite being implemented in Python instead of C++. 
We summarize the main contributions as follows:
\begin{itemize}
    \item We present the \tool{} framework that bridges the gap between deep learning and program language for synthesizing programs to extract semantic entity relations from visually rich documents.
    \item We propose a novel domain-specific language (DSL) and a novel synthesis algorithm that efficiently navigates the DSL's program space by pruning and sketching, in which sketches are obtained by mining patterns from training data and pruning is achieved by equivalence reduction and combinations of negative and mutually-exclusive programs.
    \item We perform extensive evaluation and show that \tool{}'s synthesized programs outperform the pre-trained language model while being more efficient in terms of memory footprints required.
\end{itemize}

\section{Related Works}\label{sec:related_works}
To automatically extract information from documents with varied layouts, there are three main directions: rule-based approach, deep learning-based approach, and program synthesis.

\noindent\textbf{Rule-based and template-based approach} Traditional approaches in document image processing makes use of heuristic rules~\cite{doctemplate_comp_blocks_HanchuanPeng2000, Bukhari2009, Bukhari2011, Ahn2017}. 
These rules rely on carefully designed feature engineering, e.g., projection profile alignments~\cite{doctemplate_comp_blocks_HanchuanPeng2000}, similar words and box size and distances~\cite{Ahn2017, Bukhari2011} to identify the desired document entities~\cite{ciravegna2001adaptive}.
Based on these alignments, researchers~\cite{yujian2007normalized, Bukhari2011} have shown that a template can also be built for each document form~\cite{field_extraction_document_images_Rusinol2013}. After building the template, visual and textual features can be used to calculate the similarity between a given template and a target document image, e.g., \cite{chum2008near, yujian2007normalized}. 
Additionally, \cite{d2018field} proposes a finer-grained template using the combinations of TF-IDF, angle, and distance between entities.
While the rule-based and template-based methods perform well for a fixed set of document formats, when there exist more varied document formats from different business vendors, the management of these templates can become intractable.
Additionally, template-based approaches are also prone to varying visual quality of documents, i.e., document images with low visual quality may unduly be classified as new templates, rendering template-based approaches imprecise. 
We aim to automatically aid developers in developing information extraction rules by automatically synthesizing programs based on example input document annotation. The synthesized rules can be seen as a matching template for each entity.

\noindent\textbf{Deep Learning-based Information Extraction} More recently, deep learning has also been applied for information extraction from visually rich documents. 
\cite{Palm2018} uses a recurrent neural network (RNN) to extract entities based on their textual features. They ignored the contribution of image and layout features and relied on the left-to-right reading order. \cite{Katti2018} proposed a new chargrid representation and used a convolutional neural network (CNN) to segment the regions containing desired entities in the chargrid. Nevertheless, 2D chargrid or segmentation-based IE methods usually do not work in the case of dense documents with little space between entities. 
Graph Convolution Network (GCN) \cite{kipf2016semi, hamilton2017inductive,dat2019towardsgnn} have also been applied for visually-rich document information extraction by converting the input document into a document graph \cite{Peng2017, song2018n, Liu2019, Qasim, Qasim2019, Wang2019} and uses graph neural networks for entity recognition by formulating the problem into node classification, in which each entity is a node. For relation extraction, LayoutLMv2~\cite{xu2021layoutlmv2} and LayoutLMv3~\cite{huang2022layoutlmv3} are language-specific versions that work only in English. Since we focus on multi-lingual relation extraction, we use the state-of-the-art multi-lingual relation extraction-LayoutXLM~\cite{Xu2020LayoutXLMMP} as the baseline.
TPP~\cite{zhang2023reading} re-formulate entity linking as token reading order and utilizes transformers to predict these reading orders.  While TPP is a promising method for entity linking, it requires additional annotations, which are not available and are not consistent with existing baselines and datasets.
Our work relies on the output of semantic entity recognition and the document graph used by Qasim et al.~\cite{Qasim2019}. However, we propose a new DSL along with a synthesis algorithm for this domain.

\noindent\textbf{Program Synthesis for Information Extraction} Program synthesis has long been used for different scenarios, ranging from programs handling primitive operations~\cite{lezama2008sketch, Nye2019a}, extracting information from spreadsheets~\cite{Gulwani2011FlashFill, Gulwani2017, Cambronero2023}, sequences~\cite{Iyer}, web data, PDF documents~\cite{mohamad2020, Sunder2019, Parthasarathy2022lrsyn} and manipulating images~\cite{barnaby2023imageeye}.
These works generally propose either a new domain-specific language or a new technique for efficiently navigating the search space in synthesizing programs.  The closest works to ours are that of ~\cite{Sunder2019, Iyer, Parthasarathy2022lrsyn} and ~\cite{barnaby2023imageeye}.
~\cite{Sunder2019} and ~\cite{Parthasarathy2022lrsyn} propose the usage of a transition system to extract the desired entity from a starting entity that matches specific conditions such as containing a word or lying on the left or right of a specific block.
~\cite{barnaby2023imageeye} proposed using a new DSL that involves the relation between entities to localize the desired image regions.
Our domain-specific language also contains elements specifying the spatial relation between entities as well as textual constraint (containing specific text or specific character) like~\cite{Sunder2019, Parthasarathy2022lrsyn}. However, we also allow a more flexible scheme in specifying structural constraints between elements. Particularly, while ~\cite{Sunder2019} and ~\cite{Parthasarathy2022lrsyn} only allow specifying a path, we allow specifying a graph as a relation constraint between entities. Our DSL also allows interaction between programs by union and exclusion of synthesized programs.

\section{Problem Definition}
\label{sec:background}

We give a brief definition of a symbolic representation of a document $D_V$ below, following Liu et al.~\cite{Liu2019}.
\begin{definition}[Symbolic Visually Rich Document)]\label{def:svrd}  A document $D_V$ is conceptualized as an ensemble of entities denoted by $D_V = \{e_1, e_2, \ldots e_N\}$, where $N$ is the total number of entities in the documents. Each entity $e_i$ is a tuple of $(e_i.\textbf{text}, e_i.\textbf{pos}, e_i.\textbf{lbl})$ where $e_i.\textbf{text}$ is the textual content of the entity, $e_i.\textbf{pos}$ is characterized by coordinates $(e_i.\textbf{x}_0, e_i.\textbf{y}_0, e_i.\textbf{x}_1, e_i.\textbf{y}_1)$ denotes the rectangular bounds of the entity, originating from the starting coordinate $(e_i.\textbf{x}_0, e_i.\textbf{y}_0)$ to its conclusive boundary at $(e_i.\textbf{x}_1, e_i.\textbf{y}_1)$ and $e_i.\textbf{lbl}$ is the predicted label of each entity from semantic entity recognition task.
\end{definition}

To facilitate the ground for synthesizing relational rules on these visually rich documents, we use $e_i.\textbf{center}$ to denote the center $(\frac{e_i.\textbf{x}_0 + e_i.\textbf{x}_1}{2}, \frac{e_i.\textbf{y}_0 + e_i.\textbf{y}_1}{2})$. We also build a set of relations $D_R = \{(r_{ij})\}$ following Qasim~\cite{Qasim2019}. Each relation $r_{ij}$ is a directed spatial relation between two entities $e_i = r_{ij}\textbf{.from}$ and $e_j = r_{ij}.\textbf{to}$ in $D_V$ along with a relation type $r_{ij}.\textbf{lbl}$ that is either \textbf{top, down, left, right}, denoting the spatial alignments between entities.

Given this definition of symbolic visually rich document, we define the task of semantic entity linking in Definition~\ref{def:sem_linking}.

\begin{definition}[Semantic Entity Linking] \label{def:sem_linking}  Given a Symbolic VRD \( D_V \) as defined in Definition~\ref{def:svrd}, the Semantic Entity Linking outputs a set of connections, \( \mathcal{L} \), between pairs of elements \( (e_i, e_j) \in D_V \times D_V \) such that $e_j$ has semantic connections (i.e., is key, value or subheader of) $e_i$.
\end{definition}


In the context of visually rich documents, the ``connection'' in semantic entity linking refers to the relational information embedded within the document elements. These connections are not spatial or sequential linkages but are semantic associations that define the hierarchy of information in the document.

For instance, consider a form in Figure~\ref{fig:motiv_example}: the two entities ``Event Name'' and ``The USQ Spring Festival'' have a semantic association:  The former is the question (i.e., key) while the latter is the answer. Similarly, in the same form, a section titled ``Special Event Information Sheet'' has associations with its sub-sections (e.g., ``General Information'' and ``Hours''), which are semantically linked as they elaborate on the broader ``Special Event Information Sheet'' category. The problem of identifying these relationships is called semantic entity linking.

To formulate the input and output of program synthesis, we present a formalization for semantic entity linking in the context of program synthesis. Suppose we have a dataset consisting of instances of symbolic visually rich documents, each represented as $D_i$ where $D_i = \{e_1, e_2, \ldots e_N\}$ are elements defined in Definition~\ref{def:svrd}, alongside with these documents, we are provided with the semantic link between $L_i$ existing elements, where each semantic link $(e_j, e_k) \in L_i$ is an ordered pair of entities from $D_i$, indicating a directional relationship from one element to another (i.e., $L_i \subseteq D_i\times D_i$).
Semantic entity linking aims to a synthesize program $P$, that can accurately identify the relationships between different elements within these documents. The program synthesis task can be formally expressed as follows:
\begin{equation}
\text{ Find } P \text{ s.t. } \bigwedge\left\{
\begin{array}{l}\forall ((e_1, e_2) \in L_i), (e_1, e_2) \in P(D_i) \\
\forall ((e_1, e_2) \not \in L_i), (e_1, e_2) \not\in P(D_i) \\
\end{array}\right.
\end{equation}
Each synthesized program $P$ is a specific program that can output the semantically linkable pairs, given a document $D_i$ as input.

In actual implementation, we encode the specifications in the form of a list of pairs that are linked towards each other $M_i = \{(e_1, e_2)\}$. $M_i$ would consist of all pairs of question-answer or header-question/header-sub headers.
Note that the specifications only made use of the symbolic visually rich documents, while the relation between elements in the document will later be used for the domain-specific language programs.



\section{Domain Specific Language}\label{sec:dsl}
In this section, we explain in detail our proposed DSL - a context-free grammar - that is specifically used to describe the search space of the program synthesis task. In the domain of symbolic visually-rich documents, our proposed grammar is engineered to be sufficiently expressive to enable sophisticated operations and queries while maintaining the tractability of the search space.



\subsection{Grammar}
We propose a DSL in Figure~\ref{fig:dsl} and describe the syntax of the DSL below.

\textit{Programs and Sub-programs($P$)}: At the highest level, programs are either \textbf{Find}, \textbf{Union}, or \textbf{Exclude} programs. To describe in a bottom-up manner, each \textbf{Find} program can be regarded as a single query that returns linkable pairs. All three types of programs return linkable pairs. \textbf{Union} programs aggregate results from multiple sub-programs, and \textbf{Exclude} programs filter out results of a specific program (first argument) from another program (second argument).

{\captionsetup{skip=2pt} 
\begin{figure}
\begin{align*}
\begin{array}{lcl}
P & \coloneq & \emptyset \,|\, \textbf{Union}(\{P\}) \,|\, \textbf{Exclude}(P, P) \,|\,\\
                  & & \textbf{Find}(\{ V_{set} \}, \{ R_{set} \}, C , \{ V_{set} \})\\
 C & \coloneq & \textbf{And}(C, C) \,|\, \textbf{Not}(C) \,|\, PC \\
PC & ::= &  \textbf{Rel}(V, R, V) \,|\,  LblC \,|\, StrC  \,|\, FloatC \,|\, \textbf{True} \,|\, \textbf{False} \\
LblC & ::= & VLabel == VLabel \,|\, RLabel == RLabel \\ 
StrC & ::= & \textbf{Eq}(Str, Str) \,|\, \textbf{Contain}(Str, Str) \\ 
FloatC & ::= & F < F \,|\, F > F \\
VLbl & ::= & V\textbf{.label} \,|\, \textbf{Header} \,|\, \textbf{Question} \,|\, \textbf{Answer} \\ 
RLbl & ::= & R\textbf{.label} \,|\, \textbf{Left} \,|\, \textbf{Right} \,|\, \textbf{Top} \,|\, \textbf{Down} \\ 
Str & ::= & V.\textbf{text} \,|\, \textbf{``.''} \,|\, \textbf{``/''} \,|\, \textbf{``:''} \,|\, \textbf{``-''}\\ 
F  & ::= & RFloatProp \,|\, VFloatProp \,|\, FConst\\
RFloatProp & ::= & R.\textbf{mag} \\
VFloatProp & ::= & V.\textbf{x0} \,|\, V.\textbf{x1} \,|\, V.\textbf{y0} \,|\,  V.\textbf{y1} \\
V_{set}  & ::= & V \,|\, V, V_{set} \\
R_{set}  & ::= & R \,|\, R, R_{set} \\
V &  ::=  & v_0 \,|\, v_1 \,|\, \dots \,|\, v_{10} \\
R &  ::=  & r_0 \,|\, r_2 \,|\, \dots \,|\, r_{10} \\
F & ::= & \mathbf{0} \,|\, \mathbf{0.1} \,|\, \mathbf{0.2} \,|\, \dots \,|\, \mathbf{1.0} \\
\end{array}
\end{align*}
\caption{\tool{}'s proposed DSL}
\label{fig:dsl}
\end{figure}
}
As an atomic program among these three, each \textbf{Find} program takes as input 4 arguments: The list of symbolic entity variables (i.e., first $V_{set}$), the list of relation variables (i.e., the first $R_{set}$), the condition to filter these $C$, and the return set (i.e., the second $V_{set}$ or we would call $V_{set\_return}$).
Given sets of potential variables set $V_{set}$ and relation variable set $R_{set}$ define the space of possible entities and relations that can be queried: each variable $v \in V_{set}$ can be bound to any entity and each relation variable $r \in R_{set}$ can be bound to any relation. Among all the possible entities and relations in the input document, \textbf{Find} programs filter out the entities and relations that do not match the input condition $C$.
Finally, in the matched set of relations and entities, \textbf{Find} programs return linkable pairs. These pairs are constructed by taking the binding of $v_0$ (a special variable that is always in $V_{set}$) and all entities that can be bound to any variable in $V_{set\_return}$. Here is a stepwise breakdown of the process:
\begin{itemize} 
\item \textbf{Binding Exploration} Initially, the \textbf{Find} program performs a comprehensive search where it explores all possible bindings of elements from $V_{set}$ and relationships from $R_{set}$ within the entire space of the document. The result of binding exploration is the set of all possible binding $B$.
\item \textbf{Condition Verification} Among the found set, \textbf{Find} filters the unique cases where the condition $C$ holds true.
\item \textbf{Resulting Set Accumulation} Finally, among these bindings \textbf{Find} collects the valuations of variables that belong to $V_{set\_return}$.
\end{itemize}
Thus, \textbf{Find} programs return a set of linkable pairs.

Conditions $C$ are recursively defined, grounding on base cases ($PC$) and complex constructs involving logical operations (\textbf{And}, \textbf{Not}) and property comparisons within elements and their relationships, such as \textbf{Rel} for relationship properties, or direct comparisons of labels, strings, and numerical attributes using $LblC$, $StrC$, and $FloatC$, respectively. 

The label constraints $LblC$ make use of either the predicted word labels or relation labels to filter the possible valuation of entities $V$ and relations $R$. These constraints dictate which label each binding entity should have.

The float constraints $FloatC$ perform comparisons between float properties of entities and relations. These constraints are mainly used to incorporate layout features into the synthesized programs.

The string constraints, $StrC$, focus on the textual content within the visually rich documents. These constraints allow for direct comparison of text strings, encompassing operations like checking equality or verifying if a particular substring is contained within an element's text. This part of the DSL is particularly crucial for searching for elements containing certain keywords, phrases, or syntactic patterns.

The $V$ and $R$ variables represent entities and relationships in the document, respectively. Each entity can be associated with various properties, including labels ($VLbl$), textual content ($Str$), and numerical attributes ($F$), enabling a wide range of queries and operations based on these characteristics. Relationships, denoted by $R$, describe the connections or relative orientations between different entities, providing crucial context for understanding the document's structure and semantics.

The DSL also specifies sets of these entities and relationships ($V_{set}$ and $R_{set}$), allowing operations to be performed over groups of items. These sets enable batch processing of elements, such as selecting multiple entities that meet certain criteria or applying transformations to several related elements simultaneously.

\vspace{1mm}

\noindent\textbf{Extension of \tool{}'s grammar for handling tables} \dat{One of the key advantages in program synthesis is the flexibility to leverage domain-knowledge and tools via an extension of the DSL. We note that datasets used in our experiments contain tables that have varying forms. To deal with this, we extended \tool{} by using a domain-specific module, namely, the table recognition module from TATR~\cite{smock2021tabletransformer}. We name this variant $\tool{}_{Table}$. This extension can easily be adopted by extending one grammar rule from the DSL from Figure~\ref{fig:dsl}:
\begin{equation}\label{eq:rlbl_extended}
    RLbl ::= R\textbf{.label} \,|\, \textbf{Left} \,|\, \textbf{Right} \,|\, \textbf{Top} \,|\, \textbf{Down} \,|\, \textbf{Row} \,|\, \textbf{Col} 
\end{equation}
Here, \textbf{Row} and \textbf{Col} are literals denoting the same row and same column relations. We construct these relations by using the output of TATR~\cite{smock2021tabletransformer}.}

\subsection{Denotational Semantics of the DSL}
In this Section, we describe the denotational semantics of our DSL. We first describe the semantic domains of the DSL, followed by describing each evaluation rule in a bottom-up manner.

\noindent\textbf{Semantic Domains} The denotational semantics of the DSL uses the following domains: (1) $D_P$ is the domain of all possible programs, (2) $D_C$ is the domain of conditions (3) $D_V$ is the domain of visual elements in a document, (4) $D_R$ is the domain of relationships between visual elements, (5) $D_{Set}$ is the domain of sets of visual elements and (6) $D_{Bool}$ is the domain of boolean values \{\texttt{true}, \texttt{false}\}.

\noindent\textbf{Semantic Rules for Programs} The interpretation of programs is defined recursively based on the program's structure. Below we denote the semantics of the DSL programs.
The semantic of \textbf{Find} relies on evaluating the potential bindings of (e.g., possible elements that can be assigned to) each relation and variable. Thus, we define \textit{valuation function pair} $\langle b_v, b_r \rangle$ as a function that assigns values to variables and relationships:
\begin{equation}
    b_v: V \to D_V \quad b_r: R \to D_R
\end{equation}

We denote the valuation of $v$ given binding $b$ as $[\![v]\!]_b$ and valuation of $r$ given binding $b$ as $[\![r]\!]_b$: 
\begin{align}
    & [\![V]\!]_b = b_v(V) & [\![R]\!]_b = b_r(R)\\
    & [\![V.\textbf{text}]\!]_b = [\![V]\!]_b.\textbf{text} & [\![V.\textbf{lbl}]\!]_b = [\![V]\!]_b.\textbf{lbl} \\
    & [\![V.\textbf{x}_0]\!]_b = [\![V]\!]_b.\textbf{x}_0 & [\![V.\textbf{x}_1]\!]_b = [\![V]\!]_b.\textbf{x}_1\\
    & [\![V.\textbf{y}_0]\!]_b = [\![V]\!]_b.\textbf{y}_0 & [\![V.\textbf{y}_1]\!]_b = [\![V]\!]_b.\textbf{y}_1 \\
    & [\![R.\textbf{lbl}]\!]_b = [\![R]\!]_b.\textbf{lbl} & \\
    &[\![R.\textbf{mag}]\!]_b = \left\Vert [\![R]\!]_b.\textbf{from}.\textbf{center} - [\![R]\!]_b.\textbf{to}.\textbf{center}  \right\Vert_1 \span 
\end{align}
Where $\Vert \cdot \Vert$ is the $L_1$ norm between two positions. The semantic function $[\![\cdot]\!]_b$ maps syntactic constructs (e.g., conditions) to their semantic (e.g., truth values) under a binding $b$. For a condition $C$, $[\![C]\!]_b$ is \texttt{true} or \texttt{false} depending on whether $C$ holds under $b$. We describe the semantics of evaluating conditions in a bottom-up manner below:
\begin{align}
[\![ \textbf{And}(C_1, C_2) ]\!]_b &= [\![C_1]\!]_b \land [\![C_2]\!]_b\\
[\![ \textbf{Not}(C) ]\!]_b &= \textbf{not}([\![C]\!]]_b)\\ 
[\![ \textbf{Rel}(V_1, R, V_2) ]\!]_b &= ([\![V_1]\!]_{b}, [\![V_2]\!]_{b}) == [\![R]\!]_{b} \\ 
[\![ VLabel_1 == VLabel_2 ]\!]_b &= ([\![VLabel_1]\!]_b == [\![VLabel_2]\!]_b)\\ 
[\![ RLabel_1 == RLabel_2 ]\!]_b &= ([\![RLabel_1]\!]_b == [\![RLabel_2]\!]_b)\\ 
[\![ F_1 < F_2]\!]_b &= ([\![F_1]\!]_b < \![F_2]\!]_b) \\ 
[\![ Str_1 == Str_2 ]\!]_b &= [\![Str_1]\!]_b == [\![Str_2]\!]_b \\ 
[\![ \textbf{Contains}(Str_1, Str_2) ]_b &= [\![Str_1]\!]_b \supseteq [\![Str_2]\!]_b
\end{align}

Finally, let $B$ be the obtained set of all possible bindings from the binding exploration step, where a single binding $b = \langle b_v, b_r \rangle $ is a function pair that assigns every variable in $V$ or relation in $R$ to a corresponding value or relation, such that for each $v \in V$ and $r \in R$, $b_v(v) \in D_V$ and $b_r(r) \in D_R$. Given that the inner components of the \textbf{Find} programs are defined, the semantic of \textbf{Find} program 
can be defined over this set of all possible bindings $B$: 
\begin{equation}
    [\![ \textbf{Find}(V_{set}, R_{set}, C, V_{set\_return}) ]\!] = \bigcup \left\{\begin{array}{l}\textbf{result}(b, C, V_{set\_return})\\ | b \in B \end{array}\right\} 
\end{equation}

Where $\textbf{result}(b)$ is the set of values computed by applying the constraint $C$ in the \textbf{Find} operation within the context of specific binding $b$:
\begin{equation}
    \textbf{result}(b, C, V_{set\_return}) = \left\{ ([\![v_0]\!]_b, [\![v]\!]_b) \left|\begin{array}{l} [\![C]\!]_b = \texttt{true} \\ v \in V_{set\_return} \end{array} \right. \right\}
\end{equation}
Intuitively, each result is a linking pair between the binding of $v_0$ and all variable $v$ in the return set $V_{set\_return}$. 
The notation $[\![C]\!]_b$ represents the evaluation of condition $C$ given $b$ as the binding defined above.
Intuitively, a \textbf{Find} program would gather all possible bindings of each variable and relations in $V_{set}$ and $R_{set}$ and identify the bindings that match the condition $C$. After obtaining the set of valid bindings, \textbf{Find} aggregates the valuation of variables belonging to the $V_{set\_return}$.

Finally, given the semantics definition of the non-recursive \textbf{Find} program, \textbf{Union} and \textbf{Exclude} programs can be recursively defined as below:
\begin{align}
& [\![ \textbf{Union}(P1, P2) ]\!] = [\![ P1 ]\!] \cup [\![ P2 ]\!]\\
& [\![ \textbf{Exclude}(P1, P2) ]\!] = [\![ P1 ]\!] \setminus [\![ P2 ]\!]
\end{align}
In which $P_1$ and $P_2$ can be either a \textbf{Find} program or recursively, \textbf{Union} and \textbf{Exclude} programs themselves.
In Section~\ref{sec:motivating_example}, we give an example of an actual program in this DSL.

\section{Motivating Example}\label{sec:motivating_example}
{\captionsetup{aboveskip=2pt, belowskip=2pt}
\begin{figure*}
    \centering
     \makebox[\textwidth]{
    \includegraphics[width=1.05\textwidth]{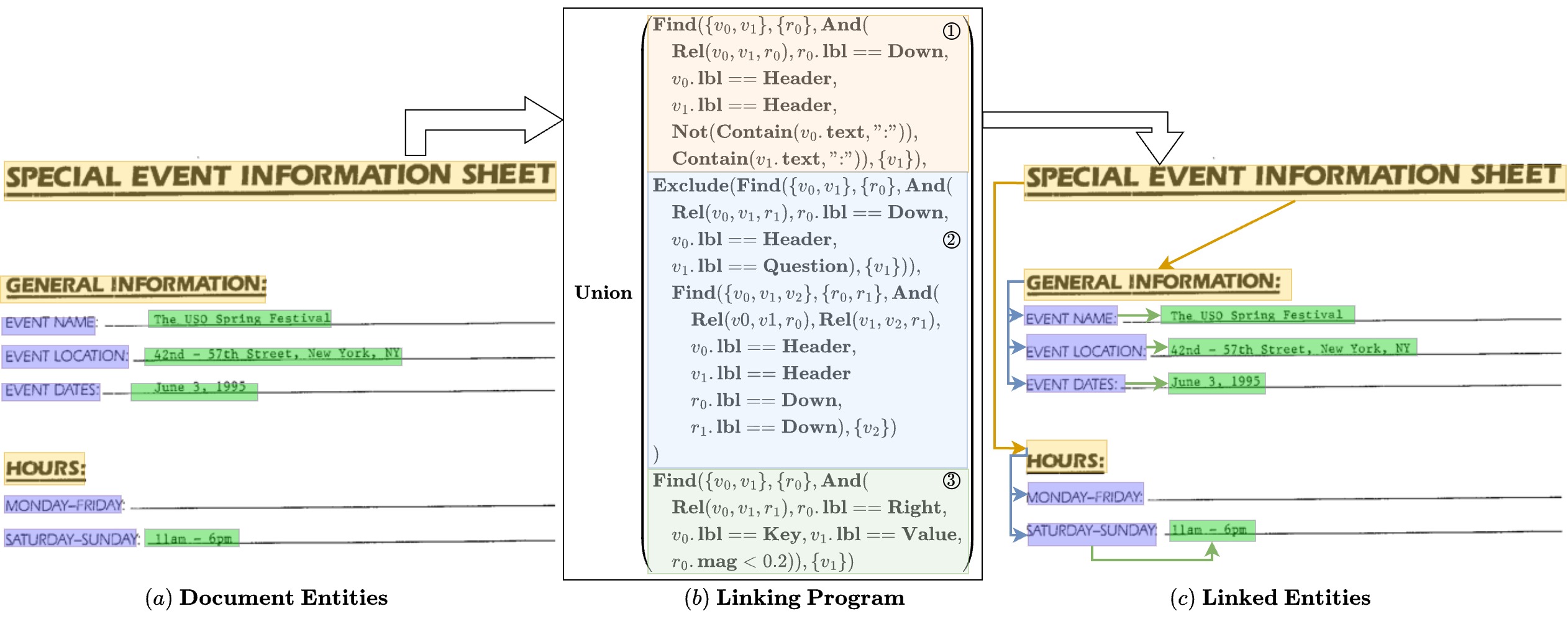}
    }
    \caption{An example of a synthesized program that links (1) header to subheader, (2) header to keys, and (3) key to value.}
    
    \label{fig:motiv_example}
\end{figure*}
}

This section presents a motivating example that uses a specific program in the DSL to link entities. Consider the event organization form in Figure~\ref{fig:motiv_example}, where each entity is assigned with a corresponding label:``Special Event Information Sheet'' is the header, ``General Information'' is also a header, and under this header there exists three keys and three values. We aim to link each header to the corresponding subheaders, each subheader to the corresponding keys, and each key to the corresponding values.

To provide a clear understanding of how one might tackle this challenge programmatically, we introduce a program depicted in Figure \ref{fig:motiv_example}b. This program links all pairs that satisfy either one of the three sub-programs. The first \textbf{Find} sub-program links between header and subheader, the second program links between header and corresponding keys, and the final program links between keys and value. The results of these programs are combined to get the final results.

All the three programs shares the variable set $\{v_0, v_1\}$ and relation variable set $\{r_0\}$. Each variable can be mapped to a corresponding entity, i.e., $e \in D_V$, and each relation variable can be mapped to corresponding relations, i.e., $(e_i,e_j) \in D_R$, in Definition 1.
Initially, without any constraint, all variables and relation variables can be bound to any entity and relations. With added constraints, such as $v_0.\textbf{lbl} == \textbf{Header}$ (variable $v_0$ has to have label $\textbf{Header}$), the space of possible binding $B$ is pruned.
In our case, the third argument of the \textbf{Find} program is the constraint to prune this search space. 
For header linking programs,
this condition prunes the possible bindings of variables and relations. In detail, it requires four sub-conditions: there must exist a relation between the valuation of $v_0$ and $v_1$ (i.e., entities that bind to variable $v_0$ and $v_1$). Furthermore, this relation has to have the label \textbf{Down}. 
The entities that can be bound to $v_0$ and $v_1$ must have label \textbf{Header}. Furthermore, $v_0$'s text must not contain a colon while $v_1$ does to avoid linking headers of the same levels.
We specifically save $v_0$ as the initial entity to be linked with. This means that for every found binding, $v_0$ is the starting entity in which we want to link every valuation of the $V_{set\_return}$ with. In this case, for every binding, we would link the valuation of $v_0$ and $v_1$. In this example, ``Special Event Information Sheet'' is linked to ``General Information'' and ``Hours''.

For header-question (header to key), we have an \textbf{Exclude} program, specifying that we want to find a set of links that satisfies the first program but avoid the links that model the second program: The first \textbf{Find} program dictates that header-linking should link two entities $v_0$, $v_1$ corresponding to \textbf{Header} and \textbf{Question} (Key) label, but avoid the keys that belong to the other headers in the program. ``General Information'' can be linked towards ``Event Name'', ``Event Location'' and ``Event Dates'' but not towards ``Monday-Friday'' and ``Saturday-Sunday'' since they already belong to another header (Hour). Since the first program's results cannot include the second program, we call the second \textbf{Find} program \textit{mutually exclusive} program of the first find program.

Finally, the third program links corresponding keys and values. Particularly, it links each key to the value on the right of it, given that the distance is sufficiently small ($r_0.\textbf{mag} < 0.2$), where $0.2$ is the ratio on the page's maximum dimension (width or height).

\section{Synthesizing Programs for VRD Information Extraction }\label{sec:synthesis}

Taking as input the set of documents $\{D\}$ along with the corresponding specifications $\{M\}$, our goal is to synthesize a single final that precisely links the entities following the specifications. For this, we first construct a relation set $D_R$ for each document $D$, in which $D_R$ consists of all relations between entities in $D$. We construct a document graph $G = \langle D, D_R \rangle$ for each document to represent the relations between entities.
Our synthesis algorithm, listed in Algorithm~\ref{alg:overall_program_synthesis}, takes these graphs as input along with the corresponding specifications and synthesizes a set of \textit{positive programs} (i.e., programs that output correct link) and \textit{negative programs} (i.e., programs that output incorrect links and should be avoided) and combine them to create the final program.

Due to the expressive power of our designed DSL, the search space for candidate programs may be large. Our synthesis algorithm effectively navigates the search space by (1) \emph{sketching} in which we start with a set of initial programs, (2) \emph{iterative refinement} via equivalence reduction with term-rewriting to avoid searching for equivalent programs, and mutually exclusive programs to increase accuracy and recall. We describe these steps in detail below.

\begin{algorithm}
\SetAlgoLined
\KwResult{Synthesized programs based on the specified grammar}
    \KwIn{Graph dataset $\{G\}$, Specs $\{M\}$, Grammar $grammar$}
    \KwOut{List of synthesized positive programs $PP$ and negative programs $NP$}

    $P \leftarrow$ \textsc{GetInitialPrograms}$(\{G\}, \{M\})$ \text{   (Algorithm~\ref{alg:initial_find_programs})}\;

    $VS \leftarrow$ \textsc{CreateVersionSpace}($P$, $\{G\}, \{M\}$)\;
    $PP, NP \leftarrow \{\}, \{\}$\;
    \For{$i = 1$ \KwTo $maxIterations$} {
        $VS, PP^*, NP^* \leftarrow$ \textsc{RefinePrograms}($VS$) (Algorithm~\ref{alg:extending_programs})\;
        $PP \leftarrow PP \cup PP^*$\, $NP \leftarrow NP \cup NP^*$\;
    }
    \Return{$PP, NP$}\;
    \caption{Overall Program Synthesis}
    \label{alg:overall_program_synthesis}
\end{algorithm}

\vspace{2mm}
\subsection{Construction of Initial Programs}\label{sec:construct_initial_program}
Recall that the search space can be large due to the expressive power of our DSL, and thus, our goal is to effectively restrict the search space by automatically constructing a set of initial programs, from which our synthesis procedure mutates and evolves to find high-quality candidate programs consistent with the provided specifications. These initial programs provide useful clues for our synthesis procedure to start the search in a more effective way.

\begin{algorithm}
\SetAlgoLined
\KwResult{List of initial find programs}
    \KwIn{List of document graph $\{G\}$ and corresponding list of specifications $\{M\}$}
    \KwOut{List of synthesized programs $P$}

    $\text{Initialize an empty list of programs }P \leftarrow \{\}$\;
    $\text{Initialize an empty list of paths }\Pi \leftarrow \{\}$\;
    \ForEach{$G_i, M_i\in\{G\}, \{M\}$}{
        \ForEach{$e_j, e_k \in M_i$}{
            $\Pi \leftarrow \Pi \cup \textsc{GetAllPaths}(G_i, e_j, e_k, 2))$\;
        }
    }
    \ForEach{$\pi \in \Pi$}{
        $V_{set}, R_{set}, Rel_{set}, V_{map} \leftarrow \{\}, \{\}, \{\}, \{\}$\;  
        $rcnt$, $vcnt$, $\textbf{v}_{last}\leftarrow 0, 0, \textbf{v}_{0}$\;  
        
        \ForEach{$(e_i, e_j) \in \pi$}{
            \If{$e_i \not \in V_{map}$}{
                $V_{map}[e_i] \leftarrow \textbf{v}_{vcnt}$\; 
                $V_{set} \leftarrow V_{set} \cup \{\textbf{v}_{vcnt}\}$, $vcnt \leftarrow vcnt + 1$\;
            }
            
            \If{$e_j \not \in V_{map}$}{
                $V_{map}[e_j] \leftarrow \textbf{v}_{vcnt}$\;
                $V_{set} \leftarrow V_{set} \cup \{\textbf{v}_{vcnt}\}$,
                $vcnt \leftarrow vcnt + 1$\;
            }
            
            $R_{set} \leftarrow R_{set} \cup \{\textbf{r}_{rcnt}\}$\;
            $Rel_{set} \leftarrow Rel_{set} \cup \{\textbf{Rel}(V_{map}[e_i], \textbf{r}_{rcnt}, V_{map}[e_j])\}$\;
            $rcnt \leftarrow rcnt + 1$,
            $\textbf{v}_{last} \leftarrow V_{map}[e_j]$\;  
        }
        
        $RelC \leftarrow \textsc{ConstructRelationConstraint}(Rel_{set})$\;
        
        $P \leftarrow P \cup \{\textbf{Find}(V_{set}, R_{set}, RelC, \{\textbf{v}_{last}\})\}$\;
    }
\caption{Constructing Initial Find Programs}
\label{alg:initial_find_programs}
\end{algorithm}

\vspace{2mm}

Our key intuition in automatically constructing the initial programs is that these programs should cover as much as possible all the links between entities specified in the specifications. To achieve this, we extract the common set of relationships from the document dataset $\{G\}$ as a set of paths between all pairs of entities and create the initial programs from these paths. Note that these paths over-approximate the relationships between entities, i.e., a path from one entity to another does not necessarily mean there is a relationship between them. Despite false positives potentially induced by these paths, initial programs constructed from the paths provide a good starting point for our synthesis algorithm to later on iteratively refine. We provide a formal definition of path in Definition~\ref{def:path}. For all collected paths, we construct the corresponding \textbf{Find} programs from these paths (See Algorithm~\ref{alg:initial_find_programs}).

\begin{definition}\label{def:path}
    Let $G = \{(D_V, D_R)\}$ be a document graph where $D_V$ is a symbolic visually rich document defined in Definition~\ref{def:svrd} and $D_R$ is the spatial relation set between $D_V$'s entities. A path $\pi$ from a starting entity $s$ to an ending entity $t$ is a sequence of $\{e_1, e_2, \ldots, e_n\}$, where $e_1 = s$ and $e_n = t$, and each consecutive pair $(e_{i}, e_{i+1})$ forms an edge in $D_R$. 
\end{definition}

After this step, we obtain the set of initial \textbf{Find} programs consisting of only variables, relational variables, and relation constraints between the variables. While these programs can cover a large amount of linking between entities, they are not precise (i.e., they might link wrong entities) since the conditions have yet to be refined. Next, based on these programs, we systematically refine programs with respect to the specifications.

\subsection{Iterative Refinement of Programs}

The initial programs can be imprecise since they lack additional structural and textual constraints. In this step, we iteratively refine the programs to find high-quality candidates. The refinement algorithm is defined in Algorithm~\ref{alg:extending_programs}.
Each iteration starts with the current set of programs and the search is performed by first converting the current condition $C$ in the \textbf{Find} program into a conjunction of $C$ and an unknown condition $\textbf{And}(C, \square)$. We enumerate possible candidates to fill in the hole while making use of equivalence reduction to avoid enumerating redundant programs. Among the enumerated candidates, we keep three types of programs: the more precise programs for the next iterations, the perfect positive programs (that output all correct pairs), and the negative programs (that output all pairs of entities that should not be linked). At the end of each iteration, we also extend the set of positive programs by combining positive, negative, and mutually exclusive programs. Below, we describe in detail each component.

\noindent\textbf{Equivalence reduction with term rewriting.} To prune the space of possible programs in the enumeration phase, we leverage equivalent reduction~\cite{smith2019equivalentreduction} in the form of term rewriting rules below. These rules are applied in a bottom-up manner to all the enumerated programs and help filter out redundant candidates.
\begin{align*}
    \textbf{And}(A, A) \leadsto A \quad 
    \textbf{And}(A, \textbf{False}) &\leadsto \textbf{False}  \quad\textbf{And}(A, \textbf{True}) \leadsto A &\\
    F < F \leadsto \textbf{False} &\quad F > F \leadsto \textbf{False}\\ 
    \textbf{Contains}(S, S) \leadsto \textbf{True} &\quad
    \textbf{Equal}(S, S) \leadsto \textbf{True} \\
    V\textbf{.label} == V\textbf{.label} \leadsto \textbf{True} &\quad R\textbf{.label} == R\textbf{.label} \leadsto \textbf{True} \\
    V\textbf{.x0} < V\textbf{.x1} \leadsto \textbf{True} &\quad V\textbf{.x1} < V\textbf{.x0} \leadsto \textbf{False} \\ 
    V\textbf{.y0} < V\textbf{.y1} \leadsto \textbf{True} &\quad V\textbf{.y1} < V\textbf{.y0} \leadsto \textbf{False}
\end{align*}
\noindent\textbf{Precision-oriented program selection.} The ultimate goal of our approach is to synthesize programs that can precisely capture the semantic relationship between entities as stated in Section~\ref{sec:background}. We formalize this goal and our detailed refinement process below. 

Given a graph dataset $\{G\}$ and a specification $\{M\}$ on this graph dataset, for each program $p$, we collect the set of all possible bindings that satisfy the condition of $p$ as $B_p$. Among all bindings in $B_p$, we further divide them into two sets $B_p^+$ and $B_p^-$.
$B_p^+$ denotes the set of bindings that \textit{models} the original specifications (i.e., $\forall b \in B_p^+, \forall \textbf{v}_{ret} \in V_{set\_return}, (b[\textbf{v}_0], b[\textbf{v}_{ret}]) \in M$), and $B_p^-$ denotes the set of bindings that do not model the specifications.
We can calculate the precision of the program by counting all the linked entities within $B_p^+$ and $B_p^-$:
\begin{equation}
    prec(B_p^+, B_p^-) = |B_p^+| / (|B_p^+| + |B_p^-|)
\end{equation}

Our observation is that, given an initial \textbf{And} constraint, any additional constraints will reduce either or both $B_p^+$ and $B_p^-$. Thus, we keep adding the constraints that improve the \textit{precision} of the program. For every \textbf{Find} program $p$, let $\{C\}$ be the set of constraints that has yet to appear in $p$ and is \textit{valid} towards $p$. 

To keep track of programs performing similarly, we keep version spaces of programs that have similar $B_p^+$ and $B_p^-$ (i.e., $VS: \langle \mathcal{P}(B), \mathcal{P}(B) \rangle \to \mathcal{P}(D_P)$), where $\mathcal{P}$ denotes power sets).
To further avoid synthesizing redundant programs, we keep track of the set of covered specifications (i.e., $\bigcup_p B_p^+$) and only accept candidates that cover a yet-covered part of specifications for further extension.

\noindent\textbf{Generating additional positive programs by using mutually exclusive programs.} While enumerating, we also keep track of perfect-positive programs (that only output correct pairs of linked entities) and perfect-negative programs (only output pairs of entities that should not be linked). At the end of each iteration, we collect additional perfect program set $PP^*$, the set of perfect negative programs $NP^*$ (programs that only give examples that do not match the specification), and the set of new version space with improved precision. These positive programs and negative programs cover the perfect set $PCover$ and $NCover$, respectively. In practice, we also observe the mutually exclusive property between certain entities. For example, in Figure~\ref{fig:motiv_example}, in the header-key linking, we can express the rule ``find the key below target header, but does not belong to another header'' as an exclude program of two \textbf{Find} programs.



\begin{algorithm}
\SetAlgoLined
\KwResult{Refined Program and Version Spaces}
    \KwIn{Version spaces $VS$}
    \KwOut{List of perfect programs $PP^*$, perfect negative programs $NP^*$ and version spaces $VS^*$}
    $Cover, PCover, NCover = \{\}, \{\}, \{\}$\;
    $VS^*, PP^*, NP^* \leftarrow \{\}, \{\}, \{\}$\;
    \ForEach{$B_p^+, B_p^-, p \in VS$}{
        \If{$B_p^+ \setminus Cover = \emptyset$}{
            \textbf{continue};
        }
        \ForEach{condition $C$ based on grammar}{
            $B_p^{+*}, B_p^{-*} \leftarrow \textsc{Filter}(B_p^+, C), \textsc{Filter}(B_p^-, C)$\;
            $p^* \leftarrow \textsc{AddConstraint}(p, C)$\;
            \If{$B_p^{+*} \setminus PCover \neq \emptyset$ and $B_p^{-*} = \emptyset$}{
                $PP^* \leftarrow PP^* \cup \{p^*\}$\;
                $PCover \leftarrow PCover \cup B_p^{+*}$\;
                $Cover \leftarrow Cover \cup B_p^{+*}$\;
                \textbf{continue}\;
            }
            \If{$B_p^{+*} \setminus Cover \neq \emptyset$ and $prec(B_p^{+*}, B_p^{-*} > prec(B_p^{+}, B_p^{-}$}{
                $Cover \leftarrow Cover \cup B_p^{+*}$\;
                $VS^*[B_p^+*, B_p^-*] \leftarrow VS^*[B_p^+*, B_p^-*] \cup \{p^*\}$\;
            }
            \If{$B_p^{+*} = \emptyset \text{ and } B_p^{-*} \setminus NCover\neq \emptyset $}{
                $NP^* \leftarrow NP^* \cup \{p^*\}$\;
                $NCover \leftarrow NCover \cup B_p^{-*}$\;
            }
        }
    }
    $EPCover \leftarrow \{\}$\;
    $p_u \leftarrow \textbf{Union}(\{PP*\} \cup \{NP*\})$\;
    \ForEach{$B_p^+, B_p^-, p \in VS$} {
        \If{$B_p^- \setminus (PCover \cup NCover)$ and $B_p^+ \setminus PCover \not \eq \emptyset$}{
            $PP^* \leftarrow PP^* \cup \textbf{Exclude}(p, p_u)$\;
            $EPCover \leftarrow EPCover \cup B_p^+$\;
        }
    }
    $PCover \leftarrow EPCover \cup PCover$\;
    \ForEach{$B_p^+, B_p^-, p \in VS^*$} {
        \If{$B_p^+ \setminus PCover = \emptyset$}{
            $VS^* \leftarrow VS^* \setminus \{(B_p^+, B_p^-, p)\}$\;
        }
    }
\caption{Refine Programs}
\label{alg:extending_programs}
\end{algorithm}
\vspace{2mm}
Finally, to construct the final program, we perform union on all of these positive sub-programs $PP$ and exclude the union of all negative sub-programs $NP$:
\begin{equation*}
    p_{final} := \textbf{Exclude}(\textbf{Union}(pp \in PP), \textbf{Union}(np \in NP))
\end{equation*}
\section{Experiment}\label{sec:exp}
We implement \tool{} in Python with over 7,000 lines of code. The experiments are performed on a computer equipped with AMD Ryzen Threadripper PRO 3995WX, with 128 GB of RAM and with RTX 4090 GPU with 24GB of VRAM. We set the maximum synthesis time of \tool{} to be 2 hours on all tasks. Since the programs can be synthesized offline, this can be seen as analogous to training time for machine learning models.

\noindent\textbf{Benchmark.}
We evaluate \tool{} on semantic entity linking task on the FUNSD~\cite{Jaume2019} and XFUND~\cite{xu-etal-2022-xfund} datasets. These two datasets contain 1,592 forms in 8 languages: English (en), German (de), French (fr), Spanish (es), Italian (it), Japanese (ja), Portuguese (pt), and Chinese (zh).
For each language, this benchmark consists of 199 scanned and annotated forms from various domains, in which $149$ forms are used for training and $50$ forms for testing.

\noindent\textbf{Baselines.} We compare \tool{} with \dat{three state-of-the-art pre-trained multilingual document understanding models: 
 LayoutXLM~\cite{Xu2020LayoutXLMMP}, InfoXLM~\cite{chi-etal-2021-infoxlm}, and XLMRoberta~\cite{conneau-etal-2020-unsupervised}}. 
\dat{These are the best-performing models in extracting question-answer relations from multilingual visually rich documents on FUNSD and XFUND benchmarks}. 
\dat{Since the original evaluation~\cite{Xu2020LayoutXLMMP} only focuses on question-answer linking, to allow these models to support the full benchmark consisting of both header-question and question-answer, we needed to extend the original data processing step to also include the header-question relations for finetuning and evaluation\footnote{As reflected in our replication package, the relevant code is found in the file \texttt{layoutlm\_re/xfund/xfund.py}, lines 225--239. The rest of the code is unchanged.}. The released models are only pre-trained models~\cite{layoutxlm-base-pretrained, infoxlm-base-pretrained, xlm-roberta-base-pretrained, infoxlm-large-pretrained, xlm-roberta-large-pretrained} (i.e., only producing the probability of each word appearing in the document). Thus, a fine-tuning process is required to use them for semantic entity linking. We fine-tuned the baselines following the reported settings in~\cite{GitHubIssue5862023} and set the number of optimization steps to 40,000. We obtained comparable results to the original benchmark (See Table~\ref{tab:performance_comparison})}.

We do not compare \tool{} with any program synthesis baselines because, to the best of our knowledge, there exists no prior work on program synthesis that can be directly applied in this domain and problem settings. 

\vspace{1mm}
\noindent\textbf{Comparisons of $\tool{}_{Table}$ with extended baselines.} \dat{We compare $\tool{}_{Table}$, which is the extension of \tool{} for handling tables (described in Section~\ref{sec:dsl}), with the extended versions of the baselines, namely $\text{InfoXLM}_{Large}$ and $\text{XLMRoberta}_{Large}$.
These large models extend their base counterparts by doubling the number of layers.} 

\vspace{1mm}
\noindent\textbf{Accuracy Evaluation.}
The dataset provides all documents with semantic relations between entities. We measure the accuracy of our approach and the baseline using precision, recall, and F1-score. We consider the identified relation to be correct if it is specified in the benchmark dataset and incorrect otherwise.

\begin{whitebox}
    \textbf{Replication Package: } To support reproducibility, we provide the open-source code, documentation, and running scripts for all experiments with aforementioned settings at \url{https://doi.org/10.5281/zenodo.10386861} 
\end{whitebox}


\subsection{RQ1: Effectiveness of \tool{} and its Extension}
{\captionsetup{aboveskip=2pt}
\begin{table*}[]
\centering 
\caption{Comparison of Performance between \tool{} and LayoutXLM, Prec., Rec. are abbreviations for precision and recall. EN, DE, ES, FR, IT, JA, PT, and ZH stand for English, German, Spanish, French, Italian, Japanese, Portuguese, and Chinese respectively. \tool{}+LayoutXLM is using \tool{} in conjunction with LayoutXLM}
\label{tab:performance_comparison} 
\small
\begin{tabular}{llccccccccc}
\toprule
\textbf{Method} & \textbf{Metrics} & \textbf{EN} & \textbf{DE} &  \textbf{ES} & \textbf{FR} & \textbf{IT} & \textbf{JA} & \textbf{PT} & \textbf{ZH} & \textbf{Avg.}\\
\midrule
& Prec & 0.712 & 0.728 & 0.705 & 0.755 & 0.738 & 0.637 & 0.752 & 0.829 & 0.732 \\
VRDSynth & Rec & 0.539 & 0.552 & 0.456 & 0.698 & 0.567 & 0.275 & 0.513 & 0.589  & 0.523 \\
& F1 & \textbf{0.614} & \textbf{0.628} & 0.554 & \textbf{0.727} & 0.641 & 0.384 & \textbf{0.61} & \textbf{0.689 }& 0.606 \\\hline
& Prec & 0.428 & 0.491 & 0.592 & 0.556 & 0.610 & 0.586 & 0.431 & 0.592 & 0.534\\
LayoutXLM & Rec  & 0.438 & 0.715 & 0.788 & 0.745 & 0.765 & 0.747 & 0.742 & 0.796 & 0.717 \\ 
& F1 & 0.433 & 0.582 & \textbf{0.676} & 0.637 & 0.679 & \textbf{0.656} & 0.545 & 0.679 & \textbf{0.610} \\
\hline
& \dat{Prec} & \dat{0.193} & \dat{0.442} & \dat{0.458} & \dat{0.414} & \dat{0.542} & \dat{0.585} & \dat{0.296} & \dat{0.559} & \dat{0.436}\\

\dat{$\text{InfoXLM}_{Base}$} & \dat{Rec} & \dat{0.150} & \dat{0.642} & \dat{0.715} & \dat{0.681} & \dat{0.681} & \dat{0.694} & \dat{0.608} & \dat{0.641} & \dat{0.602}\\
& \dat{F1} & \dat{0.169} & \dat{0.532} & \dat{0.558} & \dat{0.515} & \dat{0.604} & \dat{0.635} & \dat{0.399} & \dat{0.597} & \dat{0.501}\\
\hline
& \dat{Prec} & \dat{0.240} & \dat{0.407} & \dat{0.422} & \dat{0.367} & \dat{0.470} & \dat{0.515} & \dat{0.208} & \dat{0.476} & \dat{0.388} \\
\dat{$\text{XLMRoberta}_{Base}$} & \dat{Rec} & \dat{0.223} & \dat{0.635} & \dat{0.662} & \dat{0.673} & \dat{0.669} & \dat{0.662} & \dat{0.553} & \dat{0.638} & \dat{0.589}\\
& \dat{F1} & \dat{0.234} & \dat{0.496} & \dat{0.515} & \dat{0.475} & \dat{0.552} & \dat{0.579} & \dat{0.302} & \dat{0.515} & \dat{0.459} \\

\hline
\hline
\multirow{2}{*}{VRDSynth}& Prec & 0.473 & 0.476 & 0.554 & 0.528 & 0.563 & 0.553 & 0.431 & 0.57 & 0.519\\
\multirow{2}{*}{+LayoutXLM} & Rec & 0.689 & 0.834 & 0.867 & 0.881 & 0.863 & 0.786 & 0.843 & 0.863 & 0.829 \\ 
& F1 & 0.562 & 0.606 & \textbf{0.677} & 0.66 & \textbf{0.682} & 0.649 & 0.571 & 0.687 & \textbf{0.637}\\
\hline
& \dat{Prec} & \dat{0.735} & \dat{0.757} & \dat{0.716} & \dat{0.748} & \dat{0.740} & \dat{0.665} & \dat{0.749} & \dat{0.829} & \dat{0.742}\\
\dat{$\tool{}_{Table}$} & \dat{Rec} & \dat{0.593} & \dat{0.577} & \dat{0.449} & \dat{0.718} & \dat{0.563} & \dat{0.292} & \dat{0.511} & \dat{0.605} & \dat{0.539} \\
& \dat{F1} & \dat{\textbf{0.657}} & \dat{\textbf{0.655}} & \dat{0.552} & \dat{\textbf{0.733}} & \dat{0.640} & \dat{0.405} & \dat{\textbf{0.607}} & \dat{0.700} & \dat{0.619}\\
\hline
& \dat{Prec} & \dat{0.355} & \dat{0.615} & \dat{0.626} & \dat{0.601} & \dat{0.629} & \dat{0.582} & \dat{0.466} & \dat{0.709} & \dat{0.573} \\
\dat{$\text{InfoXLM}_{Large}$} & \dat{Rec} & \dat{0.298} & \dat{0.679} & \dat{0.737} & \dat{0.713} & \dat{0.732} & \dat{0.732} & \dat{0.636} & \dat{0.716} & \dat{0.655} \\
& \dat{F1} & \dat{0.324} & \dat{0.646} & \dat{0.677} & \dat{0.652} & \dat{0.677} & \dat{0.645} & \dat{0.537} & \dat{\textbf{0.712}} & \dat{0.609}\\
\hline
& \dat{Prec} & \dat{0.375} & \dat{0.602} & \dat{0.594} & \dat{0.540} & \dat{0.628} & \dat{0.615} & \dat{0.431} & \dat{0.686} & \dat{0.559} \\
\dat{$\text{XLM-Roberta}_{Large}$} & \dat{Rec} & \dat{0.272} & \dat{0.681} & \dat{0.743} & \dat{0.660} & \dat{0.726} & \dat{0.742} & \dat{0.622} & \dat{0.712} & \dat{0.645} \\
& \dat{F1} & \dat{0.315} & \dat{0.639} & \dat{0.661} & \dat{0.594} & \dat{0.674} & \dat{\textbf{0.673}} & \dat{0.510} & \dat{0.699} & \dat{0.596} \\


\bottomrule
\end{tabular}
\end{table*}
}


\dat{We use the DSL defined in Section~\ref{sec:dsl} for the base version of \tool{} and expand the relation label grammar rule as in equation~\ref{eq:rlbl_extended} for $\tool{}_{Table}$ for all forms across 8 languages. }
For each language, we use the training documents to synthesize a program and evaluate the synthesized program on the test set. 
\dat{For LayoutXLM~\cite{Xu2020LayoutXLMMP}, InfoXLM~\cite{chi-etal-2021-infoxlm} and XLM-Roberta~\cite{conneau-etal-2020-unsupervised}, }we use the training documents for fine-tuning the pre-trained model and evaluate the fine-tuned model on the language's test set.
Table~\ref{tab:performance_comparison} presents the precision, recall, and F1 score for both \tool{}'s synthesized program and pretrained models~\cite{Xu2020LayoutXLMMP, chi-etal-2021-infoxlm, conneau-etal-2020-unsupervised}.

\dat{Regarding the F1 score, \tool{} outperforms LayoutXLM, $\text{InfoXLM}_{Base}$ and $\text{XLMRoberta}_{Base}$ in 5, 6 and 7 out of 8 languages respectively. In all 8 languages, \tool{} achieves better precisions than the pre-trained language models. 
Among all languages, \tool{} achieved the highest F1 score (0.727) in French forms. Additionally, for English forms, \tool{} achieved 41.89\%, 263.31\% and 162.39\% higher  F1 score than LayoutXLM, $\text{InfoXLM}_{Base}$ and $\text{XLMRoberta}_{Base}$, respectively. }


\dat{
Using the table module to extend \tool{} to $\tool{}_{Table}$ further improves the performance of the synthesized programs in 5 languages: English, Germany, French, Japanese, and Chinese. Additionally, $\tool{}_{Table}$ outperforms LayoutXLM, $\text{InfoXLM}_{Base}$ and $\text{XLMRoberta}_{Base}$ in 7 out of 8 languages. In comparison with large models, $\tool{}_{Table}$ achieved better average F1 score and outperforms $\text{InfoXLM}_{Large}$ and $\text{XLMRoberta}_{Large}$ in 4 out of 8 languages.
}


\dat{We also explore a combination of the techniques that have highest precision (\tool{}) and highest recall (LayoutXLM). Particularly, we combine the outputs of \tool{} and LayoutXLM and exclude the negative programs of \tool{}, i.e., $(RE_{VRDSynth} \cup RE_{LayoutXLM}) \setminus RE_{VRDSynth\_Neg}$, where:
$RE_{LayoutXLM} = \{(e_i, e_j)\}$ indicating $e_i$ should be semantically linked to $e_j$ by LayoutXLM. $RE_{VRDSynth} = \{(e_i, e_j)\}$ indicating $e_i$ should be semantically linked to $e_j$ by VRDSynth and $RE_{VRDSynth\_Neg} = \{(e_i, e_j)\}$ indicating that $e_i$ and $e_j$ should not be linked together by VRDSynth.}

\dat{Table~\ref{tab:performance_comparison} also presents the precision, recall, and F1 score of the combined approach (\tool{} + LayoutXLM) for each language. (\tool{}+LayoutXLM) achieves higher recall and F1 but lower precision than the individual approaches.
When we combine the outputs of two methods in this manner, since the number of ground truth relations remains the same, the recall always increases given the combined true positives  (TP) from two methods. The combined precision, however, may decrease with the increased false positives (FP). As an example, in the case of Spanish (Es): VRDSynth’s TP, FP, and FN (False Negatives) relations are: 483, 202, and 576 respectively with a precision 0.705 and LayoutXLM’s TP, FP, and FN relations are: 815, 575, and 224 respectively with a precision 0.592. The combination VRDSynth+LayoutXLM’s number of TP, FP, and FN relations are 918, 740 and 141 respectively with a precision 0.554.
This means that while LayoutXLM complements VRDSynth (hence, the increased TP), it also adds more false negatives due to low precision (large FP), resulting in the dropped precision. However, given the increased recall, in terms of overall F1 score, this combination still outperforms both VRDSynth and pre-trained language model baselines, and achieved the highest average F1 score of 0.637.}
\begin{graybox}
    \textbf{RQ1 Conclusion:} \tool{}, without requiring pre-trained data or pre-training procedure, achieved significantly higher average precision than all baselines and outperformed LayoutXLM on 5 out of 8 languages in terms of F1 score. \dat{$\tool{}_{Table}$ improved \tool{}'s performance in 5 languages, outperformed all the base variants of pre-trained models in 7 out of 8 languages and large variants of pre-trained models on 4 out of 8 languages.}
\end{graybox}

\subsection{RQ2: Contribution of negative and mutually-exclusive sub-programs}
\tool{} makes use of both positive and negative sub-programs in the synthesis procedure. To measure the contribution of negative programs, we perform a comparison of synthesized programs' performance before and after adding negative/mutually exclusive programs in terms of F1 score in Figure~\ref{fig:abbl}.

{\captionsetup{skip=2pt, belowskip=-10pt}
\begin{figure}
    \includegraphics[width=\linewidth]{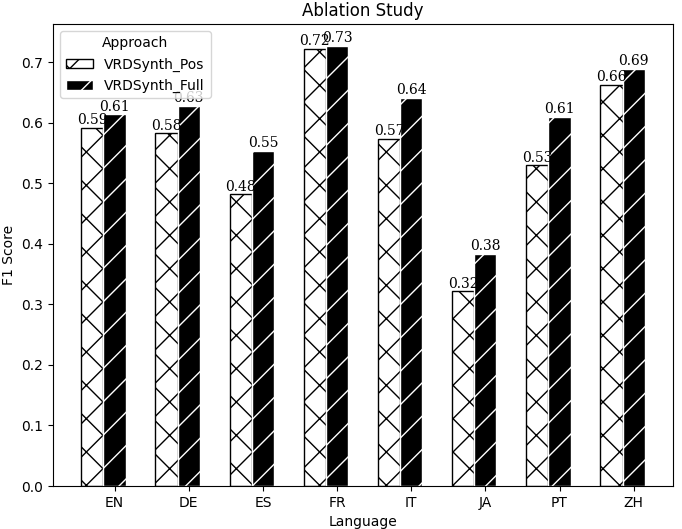}
    \caption{Comparison between \tool{} with ($\tool{}\_\text{Full}$) and without ($\tool{}\_\text{Pos}$) the usage of negative and mutually exclusive sub-programs.}
    \label{fig:abbl}
\end{figure}
}

Results show a consistent improvement of the F1 score of \tool{} in all languages. In detail, we can see improvements of 2\%, 5\%, 9\%, 1\%, 7\%, 6\%, 8\%, 3\% on en, de, es, fr, it, ja, pt, and zh respectively. This constitutes an overall improvement of 5\% in terms of F1 score.
\begin{graybox}
    \textbf{RQ2 conclusion:} Joining programs give an overall improvement of 5\% in F1 score. The highest improvement is 9\% in Spanish.
\end{graybox}
{\captionsetup{aboveskip=2pt}
\begin{table*}[]
    \centering
    \caption{Inference time comparison of \tool{}'s variants and baselines, all measurements are in seconds.} 
    \small
    \begin{tabular}{l|cc|cc|cc|cc|cc|cc}
    \hline
    \multirow{2}{*}{\textbf{Lang}} & \multicolumn{2}{c|}{$\tool{}_{full}$} & \multicolumn{2}{c|}{$\tool{}_{pos}$} & \multicolumn{2}{c|}{\dat{$\tool{}_{Table}$}} & \multicolumn{2}{c|}{LayoutXLM} & \multicolumn{2}{c|}{\dat{$\text{InfoXLM}_{Large}$}} & \multicolumn{2}{c}{\dat{$\text{XLMRoberta}_{Large}$}}  \\
    \cmidrule{2-3} \cmidrule{4-5} \cmidrule{6-7} \cmidrule{8-9} \cmidrule{10-11} \cmidrule{12-13}
     & $\textbf{Mean}$ & \textbf{Std.} & \textbf{Mean} & \textbf{Std.} & \dat{\textbf{Mean}} & \dat{\textbf{Std.}} & \textbf{Mean} & \textbf{Std.} & \dat{\textbf{Mean}} & \dat{\textbf{Std.}}& \dat{\textbf{Mean}} & \dat{\textbf{Std.}}  \\
     \hline
     En & 1.68 & 1.27 & 0.64 & 0.495 & \dat{4.49} & \dat{8.28} & 1.88 & 0.41 & \dat{4.56} & \dat{1.01} & \dat{3.07} & \dat{0.68} \\
    De & 3.71 & 2.96 & 0.86 & 0.69 & \dat{19.81} & \dat{47.53} & 2.45 & 0.85 & \dat{5.54} & \dat{2.00} & \dat{3.67} & \dat{1.33} \\
    Es & 5.23 & 2.57 & 1.27 & 0.63 & \dat{10.59} & \dat{23.65} & 2.80 & 1.02 & \dat{6.28} & \dat{2.43} & \dat{4.18} & \dat{1.63} \\
    Fr & 3.91 & 3.45 & 0.80 & 0.56 & \dat{8.13} & \dat{28.44} & 2.68 & 1.17 & \dat{6.17} & \dat{2.57} & \dat{4.13} & \dat{1.73} \\
    It & 4.99 & 3.30 & 1.12 & 0.75 & \dat{20.82} & \dat{64.92} & 3.42 & 1.33 & \dat{7.94} & \dat{3.15} & \dat{5.32} & \dat{2.12} \\
    Ja & 2.87 & 2.71 & 1.07 & 1.02 & \dat{5.77} & \dat{8.64} & 4.99 & 2.46 & \dat{6.03} & \dat{2.23} & \dat{4.03} & \dat{1.51} \\
    Pt & 5.76 & 6.23 & 1.34 & 1.42 & \dat{24.38} & \dat{47.17} & 3.20 & 1.19 & \dat{7.24} & \dat{2.56} & \dat{4.84} & \dat{1.72} \\ 
    Zh & 5.06 & 2.434 & 1.13 & 0.54 & \dat{42.61} & \dat{59.22} & 4.35 & 1.99 & \dat{5.94} & \dat{2.34} & \dat{3.93} & \dat{1.56} \\
    \hline
   \textbf{Avg.} & 4.15 & 2.95 & 1.03 & 0.76 & \dat{17.08} & \dat{35.98} & 3.22 & 1.30 & \dat{6.21} & \dat{2.29} & \dat{4.14} & \dat{1.54}\\
      \hline
     
    \end{tabular}
    \label{tab:inference_time}

\end{table*}
}
{\captionsetup{aboveskip=2pt}
\begin{table*}[]
    \centering
    \caption{Storage and memory efficiency of \tool{}'s variants and pre-trained language models}
    \small
    \begin{tabular}{l|cccccc}
    \hline
         &  $\tool{}_{full}$ &  $\tool{}_{pos}$ & \dat{$\tool{}_{Table}$} & LayoutXLM   &  \dat{$\text{InfoXLM}_{Large}$} & \dat{$\text{XLMRoberta}_{Large}$}\\ \hline
        \textbf{Storage}  & 1020 KB & 424 KB  & \dat{2.68 MB} &  1.48 GB & \dat{2.24 GB} & \dat{2.24 GB}\\
        \textbf{Memory}   & 380 MB & 378 MB  & \dat{1.20 GB} & 3.04 GB   & \dat{5.40 GB} & \dat{4.34 GB} \\
        \hline
    \end{tabular}
    \label{tab:mem_efficiency}
\end{table*}
}
\subsection{RQ3: Efficiency of \tool{}}
We measure the inference time and memory usage of \tool{}'s variants - $\tool{}_{pos}$ and $\tool{}_{full}$ (using only positive programs and all programs, respectively) along with LayoutXLM. To provide a precise comparison, we set the number of available CPU cores to 1 and do not use GPU. 

\subsubsection{RQ3.1. Inference time efficiency}
We note down the mean and standard deviation of runtimes of the synthesized programs generated by different variants of \tool{} versus the baselines in Table~\ref{tab:inference_time}. ~\tool{} inference time is 4.15 seconds on average, this speed is better than $\text{InfoXLM}_{Large}$, while being comparable with $\text{XLMRoberta}_{Large}$ and is slightly less efficient than LayoutXLM. 
The positive-program-only version of \tool{} is much more efficient, running on an average of only 1 second per document.
\dat{$\tool{}_{full}$ is more efficient than its extension, $\tool{}_{Table}$ which takes on average of 17.08 seconds for each form, due to additional same rows and columns relations detected by the table module.}

We note that \tool{}'s inference is implemented in Python, while \dat{the baselines are} based on Pytorch, which is implemented in C++, thus, in future implementation, it would be interesting to see the improvement in using C++ for executing \tool{}'s synthesized programs.

\subsubsection{RQ3.2. Memory efficiency}
Table~\ref{tab:mem_efficiency} summarizes efficiency of all techniques. \tool{} requires only 1 MB of storage and 380 MB of memory for inference, significantly lower than LayoutXLM at 1.48GB for storage and 3.04GB for memory inference. \dat{$\tool{}_{Table}$, despite requiring more memory than \tool{}, is still significantly more efficient than all pre-trained model baselines and requires only $2.68$ MB and $1.2$GB for storage and inference respectively, versus 2.24 GB and 5.40 GB of the best baseline - $\text{InfoXLM}_{Large}$}.

\begin{graybox}
    \textbf{RQ3 Conclusion:} \tool{} has an average inference time comparable with pre-trained models, despite being implemented in Python instead of C++. The positive-program-only variant of \tool{} - $\tool{}_{pos}$ runsthree times faster than LayoutXLM. \dat{The table variant $\tool{}_{Table}$, while performed the best, is less efficient than the baselines.} For memory efficiency, \dat{all variants of \tool{} are more efficient than the baselines.}
\end{graybox}
\vspace{2mm}

\section{Threats to Validity}
\noindent\textbf{Threats to internal validity} \dat{Relate to possible errors in our implementation and experiments. We have rechecked our implementation and experiments and fixed the errors that we have found. Still, there could be additional errors that we did not notice. We have carefully constructed a replication package and ensured that the exact commands and hyperparameters used for reproducing the experiments are included.}

\noindent\textbf{Threats to external validity} \dat{Correspond to the generalizability of our findings. VRDSynth consists of two components: the DSL and the synthesis algorithm. 
The DSL is designed based on various kinds of documents with various languages so it should cover various structures, still, it is possible that there exists specific documents where structures are not represented by the DSL. 
Furthermore, the synthesis algorithm works by identifying subgraphs from document graphs that match specific patterns in terms of structure and properties. These patterns are initially mined from the dataset and then iteratively extended. Thus, in the presence of long-distance and rare relations, it will be hard to either discover this relationship with mining or iterative refinement of programs. This problem is difficult for any approach handling VRDs. 
We tried mitigating the threat on the generalizability of both DSL and synthesis algorithm by experimenting on datasets containing varied real-world documents such as FUNSD and XFUND. 
To further mitigate the former threat, we also experimented with the extensibility of VRDSynth's DSL to adopt a domain-specific module to tackle table structures, namely, $\tool{}_{Table}$ without changing the synthesis algorithm, however, it is still possible that the DSL might need to be extended again to tackle even more challenging domains.}

\noindent\textbf{Threats to construct validity} \dat{Correspond to the suitability of our evaluation metrics. For RQ1 and RQ2, we measured the performance of all baselines using well-known metrics- precision, recall and F1 score. For RQ3.1, we measured both the mean and standard deviation of inference time in seconds for each language and baseline. For RQ3.2, we measured both the maximum storage and runtime memory of all baselines.}

\section{Conclusion and Future Works}\label{sec:conclusion}
We introduced a synthesis-based approach towards visually rich document understanding. In detail, we proposed a new expressive DSL enabling us to represent programs that link entities based on spatial and textual constraints. Using this DSL, we proposed a novel synthesis algorithm, ~\tool{} that efficiently synthesizes programs based on the DSL. The synthesized programs outperformed LayoutXLM, the state-of-the-art in multi-language document understanding, in 5 out of 8 languages in the joint benchmark of FUNSD and XFUND.
Our evaluation further shows that the technique of joining and mutually excluding negative programs gives a boost in terms of synthesized program effectiveness and that they are memory and computationally efficient.

In future works, we plan to leverage semantic embeddings for synthesizing more concise programs.
\bibliographystyle{unsrt}  
\bibliography{library, extra_bib}
\end{document}